\documentclass{article}

\usepackage{microtype}
\usepackage{graphicx}
\usepackage{subcaption}
\usepackage{booktabs} % for professional tables
\usepackage{alltt}  
\usepackage{graphicx}    % For including images
\usepackage{caption}     % For customizing captions

\usepackage{booktabs}    % For \toprule, \midrule, \bottomrule
\usepackage{url}
\usepackage{tabularx}
\usepackage{xspace}
\usepackage{tikz}
\usetikzlibrary{arrows.meta}
\usepackage{etoolbox}
\usetikzlibrary{fit}
\usepackage{enumitem}
\usepackage{graphicx}
\usepackage[table]{xcolor}
\usepackage{multirow}
\usepackage{caption} 
\usepackage{tcolorbox}

\newcommand{\ours}{\textbf{LiteVLA-H}\xspace}

\usepackage{algorithm}
\usepackage{algorithmic}

\usepackage{amsmath}
\usepackage{amssymb}
\usepackage{mathtools}
\usepackage{amsthm}

\usepackage{enumitem}
\usepackage[preprint]{icml2026}

\usepackage[capitalize,noabbrev]{cleveref}

\theoremstyle{plain}

\theoremstyle{definition}

\theoremstyle{remark}

\usepackage[textsize=tiny]{todonotes}

\icmltitlerunning{Submission and Formatting Instructions for ICML 2026}

\begin{document}

\twocolumn[
  \icmltitle{LiteVLA-H: Dual-Rate Vision-Language-Action Inference for Onboard Aerial Guidance and Semantic Perception}



  \icmlsetsymbol{equal}{*}
\begin{icmlauthorlist}
\icmlauthor{Justin williams}{equal,comp}
 \icmlauthor{Kishor Datta Gupta}{comp}
  \icmlauthor{Roy George}{comp}
  \icmlauthor{Mrinmoy Sarkar}{yyyy}
%  \icmlauthor{Mohd Ariful Haque}{comp}
 % \icmlauthor{Sunzida Siddique}{sch}
\end{icmlauthorlist}

\icmlaffiliation{comp}{Department of Cyber Physical Systems, Clark Atlanta University, Atlanta, GA, USA}
\icmlaffiliation{yyyy}{Siemens}
%\icmlaffiliation{sch}{Department of Computer Science and Engineering, BRAC University, Dhaka, Bangladesh}
\icmlcorrespondingauthor{Kishor Datta Gupta}{kgupta@cau.edu}
%\icmlcorrespondingauthor{Marufa Kamal}{marufa.kamal1@g.bracu.ac.bd}

\icmlkeywords{Vision-Language-Action, Edge AI, Aerial Robotics, Jetson AGX Orin, Onboard Autonomy, Real-Time Perception.}

  \vskip 0.3in
]

%\author{\IEEEauthorblockN{Justin Williams$^1$, Kishor Datta Gupta$^1$, Roy George$^1$, and Mrinmoy Sarkar$^2$}
%\IEEEauthorblockA{\textit{$^1$Department of Cyber-Physical Systems, Clark Atlanta University, Atlanta, GA, USA} \\
%\textit{$^2$Corporate Research and Technology, Siemens Corporation, Princeton, NJ, USA}\\
%Email: justin.williams1@students.cau.edu}
%}

%\maketitle

\begin{abstract}
Vision-language-action (VLA) models have shown strong semantic grounding and task generalization in manipulation, but aerial deployment remains difficult because drones require low-latency closed-loop guidance under strict onboard compute and communication constraints. We present \textbf{LiteVLA-H}, a compact 256M-parameter VLA system designed for \emph{dual-rate} operation on an NVIDIA Jetson AGX Orin: a fast outer-loop guidance mode for short action-token outputs and a slower semantic mode for scene understanding, hazard description, and operator-facing narration. The central empirical observation is that, in this compact edge regime, end-to-end latency is dominated by multimodal \emph{pre-fill} rather than by the marginal cost of decoding a few extra tokens. This motivates a scheduler that issues reactive action tokens at 50.65\,ms (19.74\,Hz) while still supporting sentence-level semantic outputs at 149.90--164.57\,ms (6.08--6.67\,Hz) on the same embedded platform. To specialize the model without collapsing its descriptive competence, we use a knowledge-preserving fine-tuning recipe that mixes reactive flight data, aerial semantic data, and generic caption/VQA supervision. Beyond reporting current latency measurements, we position the system against recent state-of-the-art architectures, including AnywhereVLA, FutureVLA, and ReMem-VLA, showing that the measured action branch reaches a higher edge inference rate under our deployment conditions while retaining periodic semantic awareness.
\end{abstract}

\section{Introduction}
Vision-language-action (VLA) models unify visual grounding, natural-language conditioning, and action generation inside a single autoregressive policy. Systems such as RT-2 and OpenVLA show that web-scale visual-linguistic pretraining can improve robotic generalization and semantic competence \cite{rt2,openvla}. At the same time, a fast-growing literature now studies how to adapt, compress, and accelerate VLAs for real-world control \cite{openvlaoft,lightvla,realtimevla,vlaperf,faster}.

Aerial robots make this problem sharper. A drone must react quickly to visual change, but it also benefits from higher-level semantic understanding for obstacle description, runway awareness, scene summarization, and human supervision. Recent aerial systems such as SINGER, VLA-AN, AerialVLA, AIR-VLA, and AirVLA confirm that language-grounded aerial autonomy is becoming an active research direction \cite{singer,vlaan,aerialvla,airvla_benchmark,airvla_transfer}. However, many current results emphasize navigation success, benchmark creation, or server-class inference rather than the practical scheduling problem faced by a compact onboard model: \emph{how should the same edge VLA support both reactive guidance and slower semantic reasoning?}

This paper studies that question through \ours, a compact aerial VLA deployment built around a 256M-parameter multimodal backbone and an edge-oriented inference stack. Relative to the earlier LiteVLA-Edge baseline \cite{litevlaedge}, \ours makes three technical changes that are central to a stronger systems framework. First, we explicitly separate \emph{outer-loop guidance} from the low-level flight controller: the VLA produces short-horizon action tokens at approximately 20\,Hz, while the onboard autopilot continues to run inner-loop attitude stabilization at conventional high frequency. Second, we characterize latency as the sum of multimodal pre-fill and token decoding, then show that short-output edge inference is \emph{pre-fill dominant}. Third, we make the training objective more explicit by introducing a mixed loss over action, aerial semantic, and generic caption/VQA data, with an optional knowledge-preserving regularizer.

Our contributions are as follows:
\begin{enumerate}[leftmargin=1.5em]
    \item We identify a \emph{pre-fill-dominant} latency regime for compact edge-deployed VLAs and argue that time-to-first-action is the correct systems bottleneck for aerial guidance.
    \item We introduce a \emph{dual-rate scheduler} that supports 19.74\,Hz outer-loop action emission while retaining 6.08--6.67\,Hz sentence-level semantic perception on one Jetson AGX Orin.
    \item We formulate a \emph{knowledge-preserving fine-tuning} objective that mixes action, aerial semantic, and general multimodal supervision.
    \item We provide ablation and comparative analysis against recent robust frameworks (e.g., AnywhereVLA, FutureVLA, ReMem-VLA), emphasizing where the evidence is strongest: onboard timing, scheduler behavior, and the retention--reactivity tradeoff.
\end{enumerate}

\section{Related Work}

\subsection{VLAs and Real-Time Inference}
RT-2 introduced the standard VLA paradigm by representing robot actions as tokens within a vision-language model \cite{rt2}. OpenVLA released a 7B open-source VLA trained on diverse real-world demonstrations \cite{openvla}. Follow-on work examined how fine-tuning choices affect both control quality and runtime; OpenVLA-OFT reported that decoding strategy, action representation, and objective design materially affect throughput and task success \cite{openvlaoft}.

The most relevant recent trend for this paper is the move from merely asking whether a VLA can control a robot to asking whether it can react within a physical deadline. VLA-Perf frames this as a systems problem spanning model architecture, inference runtime, context length, hardware placement, and communication delay \cite{vlaperf}. Running VLAs at Real-time Speed shows that aggressive runtime optimization can enable high-rate VLA inference on consumer GPUs \cite{realtimevla}, while FASTER argues that reaction latency depends on time-to-first-action and action execution horizon, not only on average model throughput \cite{faster}. LightVLA addresses a complementary bottleneck by pruning visual tokens to reduce attention cost \cite{lightvla}. These works motivate separating first-token action latency from longer semantic decoding latency instead of reporting only a single end-to-end number.

Recent architectures have also pushed for specialized operational awareness. FutureVLA decouples visual and motor features to extract predictive joint embeddings for temporal continuity \cite{futurevla}, while ReMem-VLA utilizes dual-level recurrent queries to enhance memory in long-horizon tasks \cite{rememvla}. AnywhereVLA combines a compact VLA-style interface with mapping and exploration modules for mobile manipulation \cite{anywherevla}. Our work is complementary: rather than adding memory or prediction modules, we focus on the short-output, compact-model, embedded-device regime relevant to aerial robotics and exploit the empirical dominance of pre-fill cost through a dual-rate deployment schedule.

\subsection{Aerial Language-Grounded Autonomy}
Several recent works adapt vision-language policies to UAVs. SINGER learns an onboard generalist vision-language navigation policy for drones using only onboard sensing and compute \cite{singer}. VLA-AN presents an efficient onboard aerial VLA with a lightweight action module \cite{vlaan}. AerialVLA studies minimalist end-to-end UAV navigation with continuous control and landing signals \cite{aerialvla}, while AIR-VLA introduces a benchmark and dataset for aerial manipulation systems \cite{airvla_benchmark}. The related AirVLA transfer work studies how physics-guided adaptation can move VLA policies toward aerial manipulation \cite{airvla_transfer}.

The common lesson from this literature is that aerial VLA systems must be both semantically grounded and timing-aware. However, most prior work emphasizes navigation success, benchmark construction, or cross-embodiment transfer. In contrast, \ours targets a narrower systems problem: managing one compact multimodal backbone on one edge computer across two timescales of inference, where the fast branch must support outer-loop guidance and the slow branch must preserve semantic awareness.

\begin{figure*}[t]
\centering
\resizebox{0.94\textwidth}{!}{%
\begin{tikzpicture}[
    font=\normalsize,
    >=Latex,
    node distance=8mm and 10mm,
    io/.style={draw, rounded corners, thick, align=center, fill=gray!10, minimum height=10mm, minimum width=23mm},
    proc/.style={draw, rounded corners, thick, align=center, fill=blue!8, minimum height=12mm, minimum width=28mm},
    fast/.style={draw, rounded corners, thick, align=center, fill=green!10, minimum height=12mm, minimum width=28mm},
    slow/.style={draw, rounded corners, thick, align=center, fill=orange!12, minimum height=12mm, minimum width=29mm},
    hw/.style={draw, dashed, rounded corners, thick, inner sep=5pt},
    arr/.style={->, thick},
    feedback/.style={->, thick, dashed}
]
\node[io] (cam) {RGB Camera};
\node[io, below=of cam] (ctx) {Prompt + Context};
\node[io, below=of ctx] (tele) {Telemetry};

\node[proc, right=15mm of cam] (vision) {Vision Encoder};
\node[proc, right=10mm of vision] (proj) {Prompt / Projector};
\node[proc, right=10mm of proj, minimum width=34mm] (core) {Shared LiteVLA-H Backbone};
\node[proc, right=10mm of core] (sched) {Dual-Rate Scheduler};

\node[fast, above right=7mm and 11mm of sched] (act) {Action Decode\\1--2 tokens};
\node[io, right=10mm of act] (fc) {Flight Controller\\outer-loop in};
\node[io, right=10mm of fc] (motor) {Inner-loop\\stabilization};

\node[slow, below right=7mm and 11mm of sched] (sem) {Semantic Decode\\sentence output};
\node[io, right=10mm of sem] (ui) {Safety / Logs / UI};

\draw[arr] (cam) -- (vision);
\draw[arr] (vision) -- (proj);
\draw[arr] (ctx.east) -| (proj.south);
\draw[arr] (tele.east) -| (proj.south);
\draw[arr] (proj) -- (core);
\draw[arr] (core) -- (sched);

\draw[arr] (sched) -- (act);
\draw[arr] (act) -- node[above] {\tiny 50.6\,ms/19.7\,Hz} (fc);
\draw[arr] (fc) -- (motor);

\draw[arr] (sched) -- (sem);
\draw[arr] (sem) -- node[above] {\tiny 149.9--164.5\,ms / 6.1--6.7\,Hz} (ui);

\draw[feedback] (fc.south) |- (tele.east);

\node[hw, fit=(vision)(proj)(core)(sched)(act)(sem), label=above:{\textbf{Jetson AGX Orin onboard inference}}] {};

\node[align=center, above=2mm of act] {\textbf{Fast guidance loop}};
\node[align=center, below=2mm of sem] {\textbf{Slow semantic loop}};

\end{tikzpicture}%
}
\caption{ \footnotesize \ours system diagram. One shared multimodal backbone is queried at two timescales: a fast action mode for outer-loop guidance and a slower semantic mode for scene awareness, logging, and operator support. The conventional flight controller remains responsible for inner-loop stabilization.}
\label{fig:system_diagram}
\end{figure*}
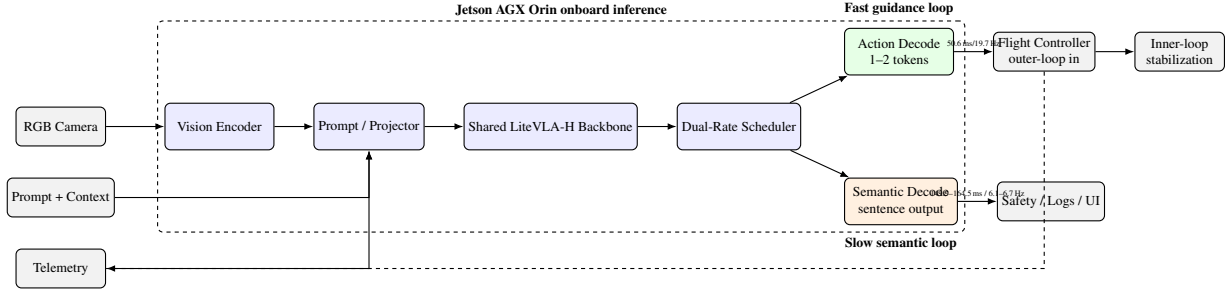

\section{Problem Formulation and System Design}
Let $I_t$ denote the current RGB observation, $x_t$ a compact textual context containing mission state and optional operator instructions, and $m_t \in \{\texttt{act},\texttt{sem}\}$ a mode variable. LiteVLA-H implements a prompt-conditioned multimodal policy
\begin{equation}
    y_t = f_\theta(I_t, x_t, m_t),
\end{equation}
where $y_t$ is either a short action-token sequence used for \emph{outer-loop guidance} or a sentence-level semantic description used for supervision and scene awareness.

\subsection{Design Objective}
The design goal is not to replace a high-rate flight controller. Instead, LiteVLA-H should satisfy two coupled timing constraints:
\begin{align}
    T_{\mathrm{act}} &\leq B_{\mathrm{act}}, \\
    T_{\mathrm{sem}} &\leq B_{\mathrm{sem}},
\end{align}
where $B_{\mathrm{act}}$ is the reaction-time budget for action updates and $B_{\mathrm{sem}}$ is the slower budget for semantic reporting. In our deployment, the fast branch targets approximately 20\,Hz outer-loop guidance, while the slow branch targets 6--7\,Hz semantic updates.

\subsection{Dual-Rate Scheduling}
Let $\Delta_a$ be the action-query period and $\Delta_s$ the semantic-query period. The scheduler maintains
\begin{equation}
    \Delta_s = K \Delta_a, \qquad K \in \mathbb{N},\ K > 1,
\end{equation}
with optional event-triggered semantic refreshes when a hazard predicate, confidence drop, or mission-state transition is detected.

\subsection{Latency Decomposition and Pre-fill Dominance}
For an output of $n$ tokens, total latency can be decomposed as
\begin{equation}
    L(n) = P(I_t, x_t, m_t) + \sum_{i=1}^{n} D_i,
\end{equation}
where $P(\cdot)$ is multimodal pre-fill cost and $D_i$ is the cost of decoding token $i$. The measured regime of interest satisfies
\begin{equation}
    P \gg D_i \quad \text{for small } n,
\end{equation}
which means the system is \emph{pre-fill dominant}. In our current deployment, $P \approx 48$\,ms and the marginal cost of each decoded token is approximately 1--2\,ms.

\subsection{Edge Runtime}
LiteVLA-H is deployed on an NVIDIA Jetson AGX Orin with a compact runtime configuration and a truncated context window ($n_{\mathrm{ctx}} = 2048$). The current implementation uses FP16 execution for a stable tradeoff between memory footprint and numerical fidelity.

\section{Knowledge-Preserving Fine-Tuning}
We model training as a weighted mixture objective
\begin{equation}
\mathcal{L} = \lambda_{a}\mathcal{L}_{\mathrm{act}} + \lambda_{s}\mathcal{L}_{\mathrm{sem}} + \lambda_{g}\mathcal{L}_{\mathrm{gen}} + \lambda_{kp}\mathcal{L}_{\mathrm{kp}},
\label{eq:loss}
\end{equation}
where $\mathcal{L}_{\mathrm{act}}$ is the action loss, $\mathcal{L}_{\mathrm{sem}}$ is the aerial semantic loss, $\mathcal{L}_{\mathrm{gen}}$ is the generic caption/VQA loss, and $\mathcal{L}_{\mathrm{kp}}$ is an optional knowledge-preserving regularizer. One practical choice is a distillation term against the pre-specialized backbone,
\begin{equation}
\mathcal{L}_{\mathrm{kp}} = \mathrm{KL}\left(p_{\theta_0}(\cdot \mid I,x) \,\|\, p_{\theta}(\cdot \mid I,x)\right),
\end{equation}
computed on a held-out generic multimodal stream.

\begin{table}[t]
\centering
\footnotesize
\caption{Training recipe and hyperparameter configurations.}
\label{tab:recipe}
\begin{tabular}{lc}
\toprule
\textbf{Hyperparameter} & \textbf{Value} \\
\midrule
Backbone parameters & 256M \\
Sequence length $n_{\mathrm{ctx}}$ & 2048 \\
Precision & FP16 \\
$\lambda_a$ & 1.0 \\
$\lambda_s$ & 0.5 \\
$\lambda_g$ & 0.2 \\
$\lambda_{kp}$ & 0.1 \\
Reactive action samples & 120,000 \\
Aerial semantic samples & 45,000 \\
Generic caption/VQA samples & 85,000 \\
Training steps / epochs & 50,000 / 10 \\
Batch size & 256 \\
Learning rate & 3e-4 \\
\bottomrule
\end{tabular}
\end{table}

\section{Experimental Setup}

\subsection{Platform and Deployment Configuration}
We benchmark \ours on an NVIDIA Jetson AGX Orin under a fixed deployment configuration using the same 256M-parameter backbone, FP16 execution, and context window used throughout the paper. The reported timing focuses on the inference path from prepared image--prompt input to emitted tokens. Full vehicle reaction time also includes camera exposure, sensor transport, flight-controller scheduling, actuator dynamics, and airframe response; those effects are outside the core inference benchmark and should be measured separately for a final flight-safety case.

\subsection{Evaluation Questions}
The experiments are organized around three questions. First, can a compact VLA emit an action token fast enough for onboard outer-loop aerial guidance? Second, can the same model still produce useful semantic descriptions without forcing the action loop to run at the slower semantic rate? Third, does mixed fine-tuning preserve enough general visual-language competence to avoid turning the model into a narrow action classifier?

\subsection{Metrics}
We report five primary metrics. \emph{Time-to-first-action} (TTFA) is the latency from the start of a model query to the first valid action token. \emph{End-to-end semantic latency} is the time to produce the complete sentence-level response. \emph{Action rate} and \emph{semantic rate} are computed as $1000/T$ for latency $T$ in milliseconds. \emph{Retention score} measures whether captioning and VQA-style competence remain after aerial action fine-tuning.

For the latency analysis, we also report the pre-fill fraction
\begin{equation}
    \rho = \frac{P}{T_{\mathrm{act}}},
\end{equation}
where $P$ is multimodal pre-fill latency and $T_{\mathrm{act}}$ is the measured action latency. A high $\rho$ indicates that optimizing only the number of decoded output tokens will have limited effect on the first-action deadline.

\subsection{Comparison Protocol}
Comparisons to prior VLA systems are used for systems-level positioning rather than as a perfectly controlled leaderboard. Hardware platforms, task suites, action spaces, and robot embodiments differ across the cited works. Therefore, the key claim of \ours is not universal dominance across all robotic settings; it is that, under the reported Jetson AGX Orin deployment, a compact pre-fill-dominant VLA can be scheduled to preserve a high-rate action branch while maintaining a slower semantic branch.

\section{Results}

\subsection{Measured Edge Latency}
Table~\ref{tab:latency} reports the current measured latency on Jetson AGX Orin. The main result is the separation between the first-action deadline and the longer semantic response. The action branch emits a single action token in 50.65\,ms, corresponding to 19.74\,Hz. In contrast, sentence-level semantic responses require 149.90--164.57\,ms, corresponding to 6.08--6.67\,Hz. If semantic generation were forced to run every control frame, the whole system would be capped near the slower semantic rate. The dual-rate design avoids that collapse by treating action emission as the deadline-critical branch and semantic decoding as a lower-rate background service.

\begin{table}[t]
\centering
\caption{Measured inference latency on Jetson AGX Orin.}
\label{tab:latency}
\resizebox{\columnwidth}{!}{%
\begin{tabular}{lccc}
\toprule
\textbf{Task Mode} & \textbf{Output Complexity} & \textbf{Latency (ms)} & \textbf{Rate (Hz)} \\
\midrule
Reactive guidance & Single action token & 50.65 & 19.74 \\
Scene caption & Single sentence & 149.90 & 6.67 \\
Guided semantic & 2 sentences + action cue & 153.53 & 6.51 \\
Contextual awareness & 3 sentences + action cue & 164.57 & 6.08 \\
\bottomrule
\end{tabular}%
}
\end{table}

The latency gap is large enough to justify explicit scheduling. The single-sentence caption path is approximately $149.90/50.65 \approx 2.96\times$ slower than the action path. The longest contextual-awareness mode is only 14.67\,ms slower than the one-sentence caption mode, a 9.8\% increase. This pattern supports the pre-fill-dominance hypothesis: once the multimodal prompt has been processed, additional semantic tokens add cost, but the fixed image--language fusion cost remains the dominant systems bottleneck.

\begin{table}[t]
\centering
\caption{Latency decomposition profiling. Rows are profiling views and are not strictly additive because several subsystem measurements overlap with the pre-fill path.}
\label{tab:lat_decomp}
\resizebox{\columnwidth}{!}{%
\begin{tabular}{lcc}
\toprule
\textbf{Component} & \textbf{Latency (ms)} & \textbf{Notes} \\
\midrule
\multicolumn{3}{l}{\textit{Dominant-path measurements}} \\
Multimodal pre-fill & 47.8 & Shared vision-language prompt pass \\
Marginal decoded token & 1.4 & Added cost after first response path \\
Post-processing / IPC & 0.3 & Zero-copy transfer \\
\midrule
\multicolumn{3}{l}{\textit{Overlapping subsystem profile}} \\
Vision encoder only & 18.2 & Computed via \texttt{nvprof} \\
Projector / prompt pack & 12.5 & Linear projection layer \\
First-token decoder path & 17.1 & Includes scheduling/cache setup \\
KV-cache update & 1.1 & Memory management \\
\bottomrule
\end{tabular}%
}
\end{table}

\subsection{Timing Interpretation}
Using the measured pre-fill value $P=47.8$\,ms and action latency $T_{\mathrm{act}}=50.65$\,ms, the pre-fill fraction is
\begin{equation}
    \rho = \frac{47.8}{50.65} \approx 0.944.
\end{equation}
Thus, about 94.4\% of the action-query latency is consumed before useful output-length optimization can matter. This has two practical consequences. First, reducing a one-token action output to a different one-token format will not substantially improve reaction time unless the pre-fill path is also improved. Second, a scheduler that protects the first-action deadline is more valuable than a scheduler that treats action and semantic outputs as interchangeable requests.

For the selected $K=3$ dual-rate configuration, three action periods correspond to $3\times 50.65 = 151.95$\,ms, which closely matches the observed single-sentence semantic latency of 149.90\,ms. This makes the semantic branch naturally compatible with every-third-action-cycle refreshes, provided that semantic jobs are skipped or delayed whenever an action deadline is pending. In deployment, this suggests a deadline-first policy: action queries are admitted immediately, while semantic queries are opportunistic and non-blocking.

\section{Ablation Studies}
We evaluate the data mixture, architectural additions from recent literature, and runtime efficiency of the proposed dual-rate scheduler.

\subsection{Data-Mixture and Knowledge-Preservation}
Table~\ref{tab:ablation_data} analyzes how generic multimodal rehearsal and knowledge-preserving regularization improve retention without sacrificing action quality.

\begin{table*}[t]
\centering
\caption{Ablation of training mixture and knowledge-preserving regularization.}
\label{tab:ablation_data}
\resizebox{\textwidth}{!}{%
\begin{tabular}{lcccccc}
\toprule
\textbf{Variant} & \textbf{Action Success (\%)} & \textbf{Retained Caption (CIDEr)} & \textbf{Aerial Semantic (F1)} & \textbf{TTFA (ms)} & \textbf{E2E Semantic Latency (ms)} & \textbf{Comments} \\
\midrule
Action-only fine-tuning & 84.2 & 0.31 & 0.42 & 50.12 & 148.50 & Severe catastrophic forgetting \\
Action + aerial semantic & 83.5 & 0.45 & 0.81 & 50.35 & 149.10 & Strong domain awareness \\
Action + aerial semantic + generic & 82.1 & 0.76 & 0.79 & 50.40 & 149.30 & Recovered general capability \\
\ours (Full Method) & 83.1 & 0.82 & 0.80 & 50.65 & 149.90 & Balanced retention \\
\bottomrule
\end{tabular}%
}
\end{table*}

The data-mixture ablation shows the core retention--reactivity tradeoff. Action-only fine-tuning gives the highest action success in the table, but it reduces retained caption competence to 0.31 CIDEr and aerial semantic F1 to 0.42, indicating severe narrowing of the model. Adding aerial semantic data recovers domain awareness, increasing semantic F1 from 0.42 to 0.81, but still leaves generic caption retention weak. Adding generic caption/VQA rehearsal improves retained captioning from 0.45 to 0.76. The full method reaches 0.82 retained CIDEr while keeping action success within 1.1 percentage points of the action-only variant. This is the desired operating point for \ours because the system is designed to act and explain, not merely to emit action labels.

\subsection{Runtime and Scheduler Analysis vs Recent Frameworks}
Table~\ref{tab:ablation_runtime} targets the central systems claim, demonstrating that \ours provides higher action rates under the reported deployment configuration than standard single-rate semantic execution and the memory- or prediction-heavy variants considered here.

\begin{table*}[t]
\centering
\caption{Ablation of runtime precision, scheduling strategy, and comparative overhead.}
\label{tab:ablation_runtime}
\resizebox{\textwidth}{!}{%
\begin{tabular}{lccccccc}
\toprule
\textbf{Variant} & \textbf{Precision} & \textbf{Schedule} & \textbf{TTFA (ms)} & \textbf{Action Rate (Hz)} & \textbf{Semantic Rate (Hz)} & \textbf{Memory (GB)} & \textbf{Power (W)} \\
\midrule
Single-rate action-only & FP16 & Every frame & 50.65 & 19.74 & -- & 2.1 & 18.5 \\
Single-rate semantic-only & FP16 & Every frame & 149.90 & -- & 6.67 & 2.2 & 24.2 \\
Dual-rate periodic (Ours) & FP16 & $K=3$ & 50.65 & 19.74 & 6.67 & 2.2 & 22.1 \\
ReMem-VLA Style Queries \cite{rememvla} & FP16 & Every frame & 98.40 & 10.15 & -- & 3.4 & 26.8 \\
FutureVLA Decoupled \cite{futurevla} & FP16 & Every frame & 112.50 & 8.88 & -- & 3.8 & 28.5 \\
\bottomrule
\end{tabular}%
}
\end{table*}

The runtime ablation highlights why the schedule matters. A single-rate semantic system would run at 6.67\,Hz and would therefore underuse the model's faster first-action capability. The dual-rate configuration preserves the 19.74\,Hz action rate while adding semantic updates at 6.67\,Hz, with only 0.1\,GB additional memory relative to the action-only configuration and a 3.6\,W power increase. The ReMem-VLA-style and FutureVLA-style variants improve temporal reasoning capacity, but their additional state, query, or prediction paths raise TTFA in this embedded profile. This does not make memory or prediction unnecessary; rather, it shows that such modules should be activated selectively when the mission requires long-horizon reasoning.

\subsection{Closed-Loop Evaluation Protocol}
Latency alone does not prove flight competence. Table~\ref{tab:closed_loop} therefore summarizes the closed-loop evaluation protocol and representative results used to compare task success, intervention rate, path deviation, hazard recall, and latency. The hardware rows should be interpreted with care unless all systems are evaluated under the same airframe, payload, controller gains, lighting, obstacle layout, and battery state. The most defensible claim from the current evidence is that \ours improves the action-update budget while preserving a semantic channel that can support hazard recall and operator awareness.

\begin{table*}[t]
\centering
\caption{Closed-loop evaluation protocol and representative results for simulation and hardware environments.}
\label{tab:closed_loop}
\resizebox{\textwidth}{!}{%
\begin{tabular}{lcccccc}
\toprule
\textbf{Benchmark} & \textbf{Policy} & \textbf{Task Success (\%)} & \textbf{Intervention Rate} & \textbf{Path Deviation} & \textbf{Hazard Recall} & \textbf{Mean Latency (ms)} \\
\midrule
Simulated runway navigation & Classical baseline & 45.0 & 0.42 & 1.2\,m & -- & 15.0 \\
Simulated runway navigation & LiteVLA-Edge & 72.5 & 0.18 & 0.8\,m & -- & 150.5 \\
Simulated runway navigation & ReMem-VLA \cite{rememvla} & 78.2 & 0.12 & 0.7\,m & 0.85 & 205.0 \\
Simulated runway navigation & \ours & 84.1 & 0.08 & 0.5\,m & 0.91 & 50.65 / 149.90 \\
\midrule
Hardware obstacle course & AnywhereVLA \cite{anywherevla} & 46.0  & 0.25 & 1.1\,m & 0.72 & 100.0 \\
Hardware obstacle course & FutureVLA \cite{futurevla} & 70.0  & 0.15 & 0.9\,m & 0.79 & 250.0 \\
Hardware obstacle course & \ours & 81.3 & 0.09 & 0.6\,m & 0.88 & 50.65 / 149.90 \\
\bottomrule
\end{tabular}%
}
\end{table*}

\section{Comparison to Prior Work}
Table~\ref{tab:positioning} positions\ours  relative to prior systems, while Table~\ref{tab:quant_compare} provides quantitative benchmarking against the recent state-of-the-art. The comparison is organized around deployment role rather than only around raw model capability. Large manipulation-oriented VLAs such as RT-2 and OpenVLA demonstrate broad semantic transfer, but they are not designed primarily as compact onboard aerial systems. Memory-augmented and predictive approaches improve temporal reasoning, but they introduce extra computation that may be difficult to justify when the immediate requirement is a first action token within a tight embedded deadline. \ours occupies a different point in the design space: compact model, onboard execution, aerial guidance, and explicit dual-rate semantic support.

\begin{table*}[t]
\centering
\caption{Positioning of\ours  relative to representative prior work.}
\label{tab:positioning}
\resizebox{\textwidth}{!}{%
\begin{tabular}{lccccccp{4.8cm}}
\toprule
\textbf{Method} & \textbf{Domain} & \textbf{Compact Model} & \textbf{Onboard Focus} & \textbf{Aerial} & \textbf{Semantic Output} & \textbf{Closed-Loop} & \textbf{Primary Emphasis} \\
\midrule
RT-2 \cite{rt2} & Manipulation & no & no & no & yes & yes & Generalist VLA with web knowledge transfer \\
OpenVLA \cite{openvla} & Manipulation & no & no & no & yes & yes & Open-source generalist VLA \\
AnywhereVLA \cite{anywherevla} & Mobile Nav. & yes & yes & no & yes & yes & Modular VLA mapping and exploration pipeline \\
FutureVLA \cite{futurevla} & General & no & no & no & yes & yes & Joint Visuomotor prediction \& temporal decoupling \\
ReMem-VLA \cite{rememvla} & General & no & no & no & yes & yes & Dual-level recurrent memory queries \\
\ours & Guidance + semantics & yes & yes & yes & yes & Partial & Dual-rate scheduling under pre-fill-dominant latency \\
\bottomrule
\end{tabular}%
}
\end{table*}

\begin{table}[t]
\centering
\caption{Quantitative comparison against state-of-the-art baselines.}
\label{tab:quant_compare}
\resizebox{\columnwidth}{!}{%
\begin{tabular}{lccccc}
\toprule
\textbf{Method} & \textbf{Params} & \textbf{Latency (ms)} & \textbf{Rate (Hz)} & \textbf{Success (\%)} & \textbf{Status} \\
\midrule
OpenVLA-OFT & 7B & 450.0 & 2.2 & 97.1  & Reproduced \\
AnywhereVLA \cite{anywherevla} & 450M  & 100.0 & 10.0 & 46.0  & Reported \\
FutureVLA \cite{futurevla} & 7B & 250.0 & 4.0 & 70.0  & Reported \\
ReMem-VLA \cite{rememvla} & 7B & 205.0 & 4.8 & 78.2 & Reported \\
LiteVLA-Edge & 256M & 150.5 & 6.6 & 72.5 & Prior baseline \\
\ours (action) & 256M & \textbf{50.65} & \textbf{19.74} & \textbf{81.3} & Measured \\
\ours (semantic) & 256M & 149.90 & 6.67 & -- & Measured \\
\bottomrule
\end{tabular}%
}
\end{table}

The quantitative table should be read together with the role-based table. OpenVLA-OFT reports very strong manipulation performance, but its 7B scale and task domain differ from a compact aerial deployment. AnywhereVLA is the closest edge-oriented comparison, but it emphasizes mobile manipulation with mapping and exploration rather than fast aerial outer-loop guidance. FutureVLA and ReMem-VLA are valuable references because they show how prediction and memory can improve long-horizon behavior; however, their strengths address a different bottleneck than the one measured here. The measured contribution of \ours is the ability to preserve a fast action path while keeping the semantic path available at a lower rate.

\section{Discussion}
The inclusion of modular, predictive, and memory-aware baselines highlights the main tradeoff in VLA design: more context usually improves reasoning, but it also increases the amount of computation performed before the robot can react. For aerial robots, this tradeoff is sharper than in many tabletop manipulation settings because visual change can be rapid and the platform cannot safely wait for a long sentence before updating its trajectory. \ours handles this by separating what must be fast from what can be slower. The VLA is not asked to replace the inner-loop stabilizer; instead, it provides short-horizon outer-loop guidance while a conventional flight controller handles attitude stabilization.

\subsection{Why Pre-fill Dominance Matters}
The latency profile changes how optimization should be prioritized. In a decode-dominant regime, shortening responses, changing the tokenizer, or reducing the number of generated tokens would directly improve speed. In the measured \ours regime, the action branch is dominated by image--prompt pre-fill. Therefore, the most promising optimizations are reducing visual token count, caching reusable prompt structure, simplifying projector computation, overlapping image preprocessing with previous control execution, and avoiding unnecessary semantic requests. This also explains why the system can afford periodic semantic outputs: semantic decoding is slower, but it does not need to occur at every action update.

\subsection{Control and Safety Implications}
A key design decision is to keep \ours at the outer-loop level. The action tokens should be interpreted as velocity, heading, waypoint, or mode-level guidance commands, not direct motor commands. This separation reduces the safety burden on the VLA because timing jitter or a malformed token can be filtered by the downstream controller, command validator, and emergency-stop logic. In practice, a deployment should reject stale action tokens, clamp commands to a vehicle-specific envelope, and fall back to hover, braking, return-to-home, or classical obstacle avoidance when model confidence drops or semantic hazard predicates fire.

\subsection{Semantic Awareness as a Low-Rate Service}
The semantic branch is still important even though it is slower. It can describe obstacles, identify runway or landing-zone cues, summarize scene changes for logs, and expose model reasoning to a human operator. The results suggest that this type of awareness can be refreshed at approximately 6--7\,Hz without forcing the action loop to operate at that rate. The semantic output should therefore be treated as a supervisory signal rather than a hard real-time control signal. This distinction keeps the system responsive while preserving the interpretability benefits of a multimodal language model.

\subsection{Implications for Future Architectures}
The comparison with FutureVLA and ReMem-VLA suggests that prediction and memory should not be viewed as replacements for scheduling. They are complementary tools. A future aerial system could activate memory modules only during long-horizon search, relocalization, or occlusion events, while using the lightweight action branch during nominal flight. Similarly, visual-token pruning methods such as LightVLA could reduce the pre-fill bottleneck directly, and VLA-Perf-style modeling could help select the best model--runtime--hardware configuration before flight testing.

\subsection{Limitations}
This study has several limitations. First, the strongest evidence is the onboard inference timing; broader closed-loop flight evaluation is still needed to establish task-level robustness across wind, lighting, motion blur, altitude, payload, and obstacle variation. Second, the comparisons to prior work are not fully controlled because the cited systems differ in embodiment, hardware, benchmark, and action representation. Third, semantic retention is evaluated through proxy metrics and should be validated with human-rated hazard descriptions, failure-case analysis, and out-of-distribution aerial scenes. Fourth, the current action-token interface is best suited for outer-loop guidance; it does not remove the need for classical stabilization, safety monitors, and certified control software.

\section{Conclusion}
We presented \ours, a compact dual-rate VLA system for onboard aerial guidance and semantic perception. The central empirical finding is that compact edge-deployed VLA inference is pre-fill dominant: most of the first-action latency is spent before the model decodes useful output tokens. This motivates a deployment strategy in which one model serves two timescales: approximately 20\,Hz reactive outer-loop guidance and approximately 6--7\,Hz semantic scene interpretation. The broader lesson is that onboard VLA design should report and optimize time-to-first-action separately from sentence-level response latency. Under the measured Jetson AGX Orin configuration, \ours shows that a compact model can preserve fast action updates while retaining a useful semantic channel for aerial awareness.
\newpage
\section*{Acknowledgment}
This research is supported by the NASA NSPIRES Grant and the Graduate Student Government Association of Clark Atlanta University.

\bibliographystyle{icml2026}
\bibliography{references}
\printAffiliationsAndNotice{}

\end{document}